\documentclass[twocolumn]{article}
\usepackage[dvipdfmx]{graphicx}
\usepackage{arxiv}
\usepackage{authblk}

\usepackage{multirow}%
\usepackage{amsmath,amssymb,amsfonts,amsthm}
\usepackage{mathrsfs}%
\usepackage[utf8]{inputenc}
\usepackage{hyperref}
\usepackage{enumitem}

\usepackage{xcolor}%
\usepackage{textcomp}%
\usepackage{manyfoot}%
\usepackage{booktabs}%
\usepackage{algorithm}%
\usepackage{algorithmicx}%
\usepackage{algpseudocode}%
\usepackage{listings}%

\usepackage{multirow}
\usepackage{subfig}
\usepackage{array}
\usepackage{tabularx}
\usepackage{threeparttable}
\usepackage{colortbl}

\theoremstyle{thmstyleone}%
\theoremstyle{thmstyletwo}%

\theoremstyle{thmstylethree}%
\newcommand{\ccol}[2]{ \multicolumn{#1}{c}{#2}}

\newcolumntype{P}[1]{>{\centering\arraybackslash}m{#1}}

\begin{document}

\newcommand{\myPaperShortTitle}{Hallucinations in Bibliographic Recommendation}
\newcommand{\myPaperTitle}{Hallucinations in Bibliographic Recommendation: Citation Frequency as a Proxy\\for Training Data Redundancy}
\title{\myPaperTitle}
\date{}


\renewcommand\Authfont{\bfseries}
\setlength{\affilsep}{0em}
\newbox{\orcid}\sbox{\orcid}{\includegraphics[scale=0.06]{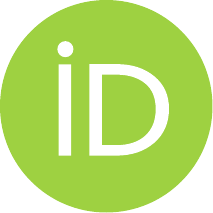} }
\author[1,2]{%
	\href{https://orcid.org/0000-0002-4618-6272}{\usebox{\orcid}\hspace{1mm}
	Junichiro Niimi\thanks{\texttt{jniimi@meijo-u.ac.jp}}
	}}
\affil[1]{Meijo University}
\affil[2]{RIKEN AIP}

\renewcommand{\shorttitle}{\myPaperShortTitle}


\twocolumn[
	\begin{@twocolumnfalse}
		\maketitle
\vspace{-3em}
\begin{abstract}
Large language models (LLMs) have been increasingly applied to a wide range of tasks, from natural language understanding to code generation. 
While they have also been used to assist in bibliographic recommendation, the hallucination of non-existent papers remains a major issue. 
Building on prior studies, this study hypothesizes that an LLM’s ability to correctly produce bibliographic information depends on whether the underlying knowledge is generated or memorized, with highly cited papers (i.e., more frequently appear in the training corpus) showing lower hallucination rates.
We therefore assume citation count as a proxy for training data redundancy (i.e., the frequency with which a given bibliographic record is repeatedly represented in the pretraining corpus) and investigate how citation frequency affects hallucinated references in LLM outputs. 
Using GPT-4.1, we generated and manually verified 100 bibliographic records across twenty computer-science domains, and measured factual consistency via cosine similarity between generated and authentic metadata. The results revealed that (i) hallucination rates vary across research domains, (ii) citation count is strongly correlated with factual accuracy, and (iii) bibliographic information becomes almost verbatimly memorized beyond approximately 1,000 citations.
These findings suggest that highly cited papers are nearly verbatimly retained in the model, indicating a threshold where generalization shifts into memorization.
\end{abstract}
\vspace{0.5em}
\vspace{2em}
	\end{@twocolumnfalse}
]

\renewcommand\thefootnote{*}
\setcounter{footnote}{0}

\section{Introduction}
Large language models (LLMs) have achieved remarkable fluency across a wide range of domains \cite{llm_sentiment_review}. 
However, they are also known to generate hallucinations that are nonsensical or unfaithful to the provided source content \cite{hallucination, hallucination_inevitable1}. 
In particular, the generation of non-existent academic references or legal precedents has been widely recognized as a critical issue \cite{hallucination2}. 
For example, in the field of marketing, where Recency–Frequency–Monetary (RFM) analysis \cite{rfm1,rfm2,rfm3} is commonly employed as a customer relationship management (CRM) \cite{loyalty}, when prompted to “Please suggest recent academic papers on RFM analysis with Author (Year) Title, Journal, Vol, No, pp style,” the model (GPT-4.1) produced the following response:
\begin{quote}
Chitturi, P., Raghunathan, B., Sciandra, R., \& Sikora, J. (2010). 
“RFM and CLV: Using Customer Data for Improved Decision Making.” 
Journal of Direct, Data, and Digital Marketing Practice, 12(1), 1–10.
\end{quote}
Although the output follows the correct bibliographic format, the paper itself does not exist. Each component imitates genuine studies, such as the author names (e.g., Chitturi and Raghunathan \cite{hallu2}), journal name (Journal of Direct, Data, and Digital Marketing Practice \cite{hallu3}), and the paper title (e.g., “RFM and CLV” \cite{hallu1}), but the numerical details are fictitious, suggesting that multiple authentic entries were probabilistically merged into a coherent yet fabricated citation.

These fabricated yet plausible references suggest that hallucinations in bibliographic recommendation may not occur arbitrarily, but rather reflect how knowledge is represented within the model. For example, the probability of reproducing training data has been shown to correlate with its frequency of appearance \cite{carlini2}. 
This study therefore focuses on bibliographic recommendation using LLMs and empirically examines how factual correctness varies with domain popularity (number of papers in the field) and citation prominence (citation count of the generated reference). 
Our findings suggest that hallucinations arise not randomly but systematically from imbalanced knowledge distributions within the representation space.

\section{Related Study}
Hallucination in LLMs has been examined from diverse perspectives \cite{hallucination,hallucination2,hallucination3,hallucination_codegen,openai_hallucinate}. 
OpenAI’s analysis \cite{openai_hallucinate} argued that reinforcement learning with human feedback (RLHF) \cite{rlhf,reinforce1} may inherently encourage hallucination, as current LLMs are penalized for responding “I don’t know” (IDK) and instead rewarded for producing statistically plausible continuations. 
This alignment objective can thus promote confident but unreliable statements.

Conversely, security-oriented studies have highlighted the opposite tendency: information repeated multiple times during pretraining is more likely to be memorized and reproduced verbatim \cite{llm_leakage1,llm_leakage2,carlini1,carlini2}. 
This view aligns with recent theoretical accounts positioning LLMs as probabilistic pattern recognizers that approximate data distributions rather than explicitly “understanding” knowledge \cite{mirchandani2023,kallens2024}. From this perspective, hallucination and {\it exposure} \cite{carlini2019} (i.e., training data leakage) represent opposite outcomes of the same probabilistic learning dynamics, where the frequency of exposure governs whether information is faithfully recalled or spuriously synthesized.

In the context of citation recommendation \cite{biblio1,biblio2,biblio3}, this implies that frequently cited papers which appear across numerous publications and other web sources are more likely to be verbatimly recalled by LLMs, whereas sparsely represented works tend to be fabricated. This study hypothesizes that hallucination in bibliographic recommendation is systematically related to {\it the training data redundancy} (i.e., the frequency with which a given bibliographic record is repeatedly represented in the pretraining corpus). 
Highly cited papers are expected to be more robustly represented, leading to lower hallucination rates, while limited-redundancy papers are more prone to plausible but non-existent references.

\section{Experiments}
\subsection{Methodology}
Bibliographic records were generated using GPT-4.1 accessed via API (knowledge cutoff: June 2024). 
To minimize domain-specific citation bias, twenty topics were selected within the field of computer science (e.g., transformer \cite{transformer}, diffusion model \cite{diffusion}, retrieval-augmented generation \cite{rag}). 
For each topic, the model was prompted to recommend five academic papers in JSON format (Fig.~\ref{fig:prompt}; see Appendix~A), yielding a total of 100 samples.

Each output record was manually validated using Google Scholar. 
Existence of the paper was confirmed primarily by its title; minor coincidences in author or journal names were not considered sufficient. 
Each record was scored as completely correct ($score=2$), partially hallucinated ($score=1$; when some metadata such as author names, journal, or year were inaccurate), or completely hallucinated ($score=0$). 
Citation counts were retrieved from Google Scholar (as of October 2025). 
Semantic similarity between generated and authentic bibliographic metadata was computed using Sentence-BERT \cite{sbert} (all-MiniLM-L6-v2) embeddings with cosine similarity, and the relationship between citation frequency and similarity was analyzed.

\subsection{Results}
\paragraph{Experiment 1.} 
We first compared low- and high-citation groups divided at the median citation count ($M_{dn} = 818$). 
A one-tailed $t$-test revealed that the high-citation group achieved significantly higher factual scores than the low-citation group ($t(98) = -5.12$, $p < .001$; $M_{\text{high}} = 1.245$, $M_{\text{low}} = 0.725$).

\begin{figure*}[htb]
   \centering
   \includegraphics[width=\linewidth]{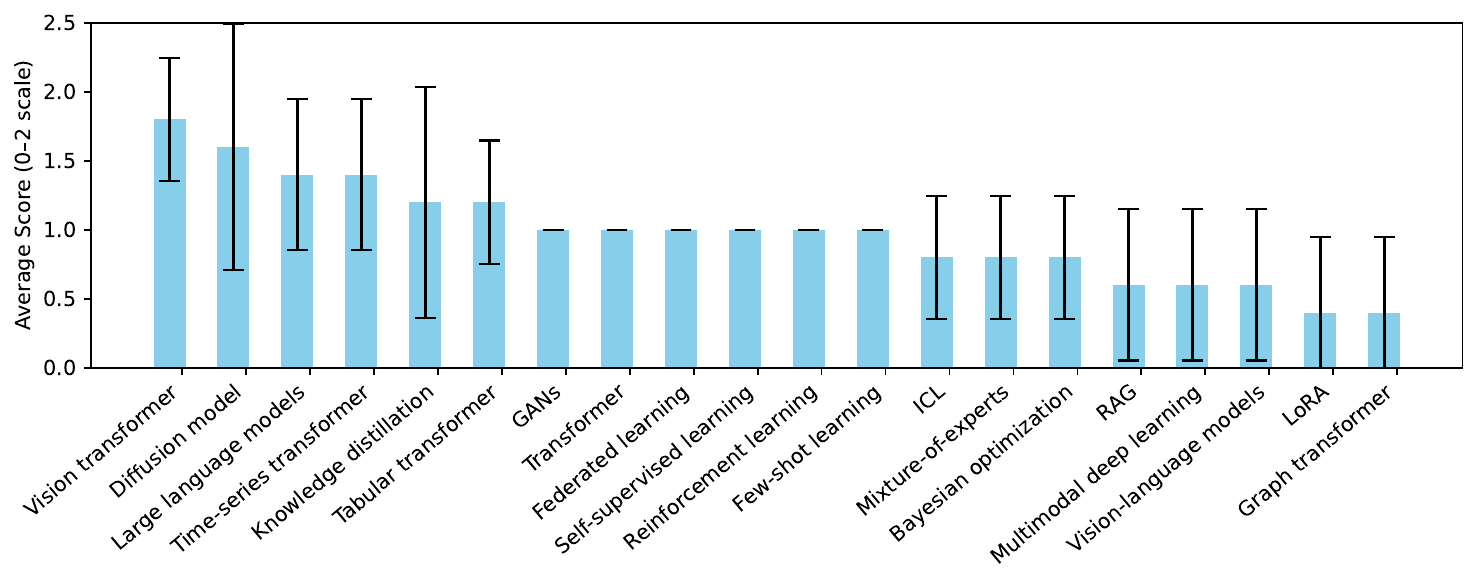}
   \caption{Average factual score by domain. Error bars indicate the 95\% confidence interval. Accuracy varies by domain familiarity.}
   \label{fig:accuracy}
\end{figure*}

\paragraph{Experiment 2.} 
Figure \ref{fig:accuracy} shows the average factual scores across research domains. The scores differ markedly among domains: Domains related to image processing, such as Vision Transformer (ViT) \cite{vit} and Diffusion Model \cite{diffusion}, achieved notably higher accuracy, whereas recent LLM-oriented techniques, such as RAG \cite{rag} and LoRA \cite{lora}, exhibited substantially lower scores. 
This discrepancy likely reflects the popularity of the research domain, which is related with the data redundancy. Consequently, hallucinations occurred more frequently in underrepresented domains such as LoRA and Graph Transformer \cite{graphtransformer}. 

\begin{figure*}[htb]
   \centering
   \includegraphics[width=0.95\linewidth]{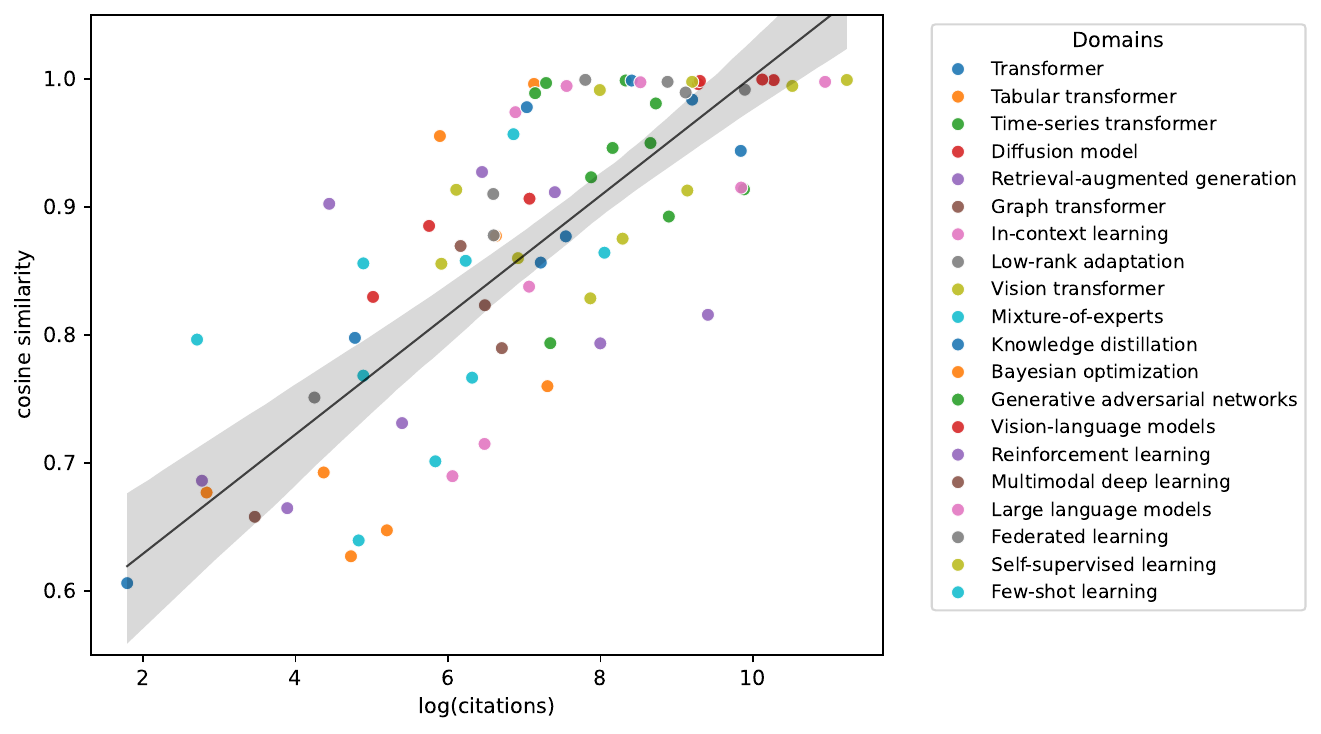}
   \caption{Relationship between citation frequency and generation fidelity. 
   Each dot represents a factual citation generated by the model, colored by research domain. 
   The dashed line indicates the fitted linear regression (95\% CI in gray). 
   The correlation ($r = 0.75$, $p < .001$) demonstrates a strong log-linear relationship between citations and factual accuracy, with saturation near $\log$(citation) $\approx$ 7.}
   \label{fig:scatter}
\end{figure*}

\paragraph{Experiment 3.} 
Using only valid records ($score > 0$; $n = 81$), we analyzed the relationship between log-transformed citation counts and cosine similarity (Fig.~\ref{fig:scatter}). 
A strong positive correlation ($r = 0.75$, $p < .001$) indicates a clear log-linear relationship, which reflects the memorization scaling law \cite{carlini2}. 
In this context, we may interpret our finding as a form of a redundancy scaling law, where the probability of factual recall significantly increases as $\log$(citation) rises.

\paragraph{Experiment 4.}
Notably, Fig. \ref{fig:scatter} also indicates the saturation near 1.0 when $\log$(citation) exceeded 7 (approximately 1,000 citations), suggesting that highly cited papers are almost verbatimly retained within the model. In that regime, frequently cited papers are not merely represented through probabilistic token associations but are instead recalled almost verbatim. 
To quantify this non-linear pattern, we conducted logistic regression using min--max-normalized cosine similarity as the dependent variable. 
The model revealed a significant positive relationship between $\log$(citation) and normalized similarity ($\beta_1 = 0.523$, $p = 0.003$) with a low intercept ($\beta_0 = -2.360$, $p = 0.020$). 
The estimated inflection point ($-\beta_0 / \beta_1 \approx 5$) corresponds to roughly 100 citations, marking the transition from generalization to memorization.

These four experiments collectively support our initial hypothesis that citation count acts as a proxy for the training data redundancy. The positive and non-linear relationship between log(citation) and cosine similarity indicates that hallucination is not random but structurally linked to uneven knowledge distributions within the model’s representation space.

\section{Conclusion}
This study empirically examined how citation frequency functions as a proxy for hallucination in bibliographic recommendation by LLMs. 
The model was instructed to output JSON-formatted results without explanations, effectively disabling
 IDK responses. In line with previous study \cite{openai_hallucinate}, such output constraints encourage the model to produce plausible yet non-existent entries. 

The key findings are as follows:  
(i) hallucination rates vary across research domains,  
(ii) citation count is strongly correlated with factual accuracy, and  
(iii) bibliographic information becomes almost verbatimly memorized beyond approximately 1,000 citations.  
In other words, while LLMs can faithfully reproduce information about highly cited papers, they struggle in domains with shorter publication histories or limited redundancy in the training corpus. 
Contrary to the discoverability phenomenon \cite{carlini2} where memorization emerges only when sufficient context is given, our results suggest the opposite direction: highly redundant knowledge can be recalled even with minimal prompting.
Although this study focused on GPT-4.1 within the computer science domain, future research should extend the analysis to other models, disciplines, and multilingual contexts to assess the generality of this threshold behavior.

\bibliographystyle{unsrt}
\bibliography{../lsys-ailet.bib}

\begin{thebibliography}{10}

\bibitem{llm_sentiment_review}
Jan~Ole Krugmann and Jochen Hartmann.
\newblock Sentiment analysis in the age of generative ai.
\newblock {\em Customer Needs and Solutions}, 11(1):3, 2024.

\bibitem{hallucination}
Sebastian Farquhar, Jannik Kossen, Lorenz Kuhn, and Yarin Gal.
\newblock Detecting hallucinations in large language models using semantic
  entropy.
\newblock {\em Nature}, 630(8017):625--630, 2024.

\bibitem{hallucination_inevitable1}
Ziwei Xu, Sanjay Jain, and Mohan Kankanhalli.
\newblock Hallucination is inevitable: An innate limitation of large language
  models.
\newblock {\em arXiv preprint arXiv:2401.11817}, 2024.

\bibitem{hallucination2}
Lei Huang, Weijiang Yu, Weitao Ma, Weihong Zhong, Zhangyin Feng, Haotian Wang,
  Qianglong Chen, Weihua Peng, Xiaocheng Feng, Bing Qin, et~al.
\newblock A survey on hallucination in large language models: Principles,
  taxonomy, challenges, and open questions.
\newblock {\em ACM Transactions on Information Systems}, 43(2):1--55, 2025.

\bibitem{rfm1}
Connie~L Bauer.
\newblock A direct mail customer purchase model.
\newblock {\em Journal of Direct Marketing}, 2(3):16--24, 1988.

\bibitem{rfm2}
Jan~Roelf Bult and Tom Wansbeek.
\newblock Optimal selection for direct mail.
\newblock {\em Marketing Science}, 14(4):378--394, 1995.

\bibitem{rfm3}
Sunil Gupta and Donald~R Lehmann.
\newblock Customer lifetime value and firm valuation.
\newblock {\em Journal of Relationship Marketing}, 5(2-3):87--110, 2006.

\bibitem{loyalty}
Jacob Jacoby and Robert~W Chestnut.
\newblock {\em Brand loyalty: Measurement and management}.
\newblock John Wiley \&amp; Sons Incorporated, 1978.

\bibitem{hallu2}
Ravindra Chitturi, Rajagopal Raghunathan, and Vijay Mahajan.
\newblock Form versus function: How the intensities of specific emotions evoked
  in functional versus hedonic trade-offs mediate product preferences.
\newblock {\em Journal of marketing research}, 44(4):702--714, 2007.

\bibitem{hallu3}
Efthymios Constantinides and Stefan~J Fountain.
\newblock Web 2.0: Conceptual foundations and marketing issues.
\newblock {\em Journal of direct, data and digital marketing practice},
  9(3):231--244, 2008.

\bibitem{hallu1}
Peter~S Fader, Bruce~GS Hardie, and Ka~Lok Lee.
\newblock Rfm and clv: Using iso-value curves for customer base analysis.
\newblock {\em Journal of marketing research}, 42(4):415--430, 2005.

\bibitem{carlini2}
Nicholas Carlini, Daphne Ippolito, Matthew Jagielski, Katherine Lee, Florian
  Tramer, and Chiyuan Zhang.
\newblock Quantifying memorization across neural language models.
\newblock In {\em The Eleventh International Conference on Learning
  Representations}, 2022.

\bibitem{hallucination3}
Hoang~Anh Dang, Vu~Tran, and Le-Minh Nguyen.
\newblock Survey and analysis of hallucinations in large language models:
  attribution to prompting strategies or model behavior.
\newblock {\em Frontiers in Artificial Intelligence}, 8:1622292, 2025.

\bibitem{hallucination_codegen}
Joseph Spracklen, Raveen Wijewickrama, AHM~Nazmus Sakib, Anindya Maiti, and
  Bimal Viswanath.
\newblock We have a package for you! a comprehensive analysis of package
  hallucinations by code generating $\{$LLMs$\}$.
\newblock In {\em 34th USENIX Security Symposium (USENIX Security 25)}, pages
  3687--3706, 2025.

\bibitem{openai_hallucinate}
Adam~Tauman Kalai, Ofir Nachum, Santosh~S Vempala, and Edwin Zhang.
\newblock Why language models hallucinate.
\newblock {\em arXiv preprint arXiv:2509.04664}, 2025.

\bibitem{rlhf}
Paul~F Christiano, Jan Leike, Tom Brown, Miljan Martic, Shane Legg, and Dario
  Amodei.
\newblock Deep reinforcement learning from human preferences.
\newblock {\em Advances in neural information processing systems}, 30, 2017.

\bibitem{reinforce1}
Daniel~M Ziegler, Nisan Stiennon, Jeffrey Wu, Tom~B Brown, Alec Radford, Dario
  Amodei, Paul Christiano, and Geoffrey Irving.
\newblock Fine-tuning language models from human preferences.
\newblock {\em arXiv preprint arXiv:1909.08593}, 2019.

\bibitem{llm_leakage1}
Katherine Lee, Daphne Ippolito, Andrew Nystrom, Chiyuan Zhang, Douglas Eck,
  Chris Callison-Burch, and Nicholas Carlini.
\newblock Deduplicating training data makes language models better.
\newblock In {\em Proceedings of the 60th Annual Meeting of the Association for
  Computational Linguistics (Volume 1: Long Papers)}, pages 8424--8445, 2022.

\bibitem{llm_leakage2}
Nikhil Kandpal, Eric Wallace, and Colin Raffel.
\newblock Deduplicating training data mitigates privacy risks in language
  models.
\newblock In {\em International Conference on Machine Learning}, pages
  10697--10707. PMLR, 2022.

\bibitem{carlini1}
Nicholas Carlini, Florian Tramer, Eric Wallace, Matthew Jagielski, Ariel
  Herbert-Voss, Katherine Lee, Adam Roberts, Tom Brown, Dawn Song, Ulfar
  Erlingsson, et~al.
\newblock Extracting training data from large language models.
\newblock In {\em 30th USENIX security symposium (USENIX Security 21)}, pages
  2633--2650, 2021.

\bibitem{mirchandani2023}
Suvir Mirchandani, Fei Xia, Pete Florence, Brian Ichter, Danny Driess,
  Montserrat~Gonzalez Arenas, Kanishka Rao, Dorsa Sadigh, and Andy Zeng.
\newblock Large language models as general pattern machines.
\newblock In {\em Conference on Robot Learning}, pages 2498--2518. PMLR, 2023.

\bibitem{kallens2024}
Pablo Contreras~Kallens and Morten~H Christiansen.
\newblock Distributional semantics: Meaning through culture and interaction.
\newblock {\em Topics in cognitive science}, 2024.

\bibitem{carlini2019}
Nicholas Carlini, Chang Liu, {\'U}lfar Erlingsson, Jernej Kos, and Dawn Song.
\newblock The secret sharer: Evaluating and testing unintended memorization in
  neural networks.
\newblock In {\em 28th USENIX security symposium (USENIX security 19)}, pages
  267--284, 2019.

\bibitem{biblio1}
Chanwoo Jeong, Sion Jang, Eunjeong Park, and Sungchul Choi.
\newblock A context-aware citation recommendation model with bert and graph
  convolutional networks.
\newblock {\em Scientometrics}, 124(3):1907--1922, 2020.

\bibitem{biblio2}
Zitong Zhang, Braja~Gopal Patra, Ashraf Yaseen, Jie Zhu, Rachit Sabharwal, Kirk
  Roberts, Tru Cao, and Hulin Wu.
\newblock Scholarly recommendation systems: a literature survey.
\newblock {\em Knowledge and Information Systems}, 65(11):4433--4478, 2023.

\bibitem{biblio3}
Jie Zhu, Braja~G Patra, and Ashraf Yaseen.
\newblock Recommender system of scholarly papers using public datasets.
\newblock {\em AMIA summits on translational science proceedings}, 2021:672,
  2021.

\bibitem{transformer}
Ashish Vaswani, Noam Shazeer, Niki Parmar, Jakob Uszkoreit, Llion Jones,
  Aidan~N Gomez, {\L}ukasz Kaiser, and Illia Polosukhin.
\newblock Attention is all you need.
\newblock {\em Advances in neural information processing systems}, 30, 2017.

\bibitem{diffusion}
Jonathan Ho, Ajay Jain, and Pieter Abbeel.
\newblock Denoising diffusion probabilistic models.
\newblock {\em Advances in neural information processing systems},
  33:6840--6851, 2020.

\bibitem{rag}
Patrick Lewis, Ethan Perez, Aleksandra Piktus, Fabio Petroni, Vladimir
  Karpukhin, Naman Goyal, Heinrich K{\"u}ttler, Mike Lewis, Wen-tau Yih, Tim
  Rockt{\"a}schel, et~al.
\newblock Retrieval-augmented generation for knowledge-intensive nlp tasks.
\newblock {\em Advances in neural information processing systems},
  33:9459--9474, 2020.

\bibitem{sbert}
Nils Reimers and Iryna Gurevych.
\newblock Sentence-bert: Sentence embeddings using siamese bert-networks.
\newblock In {\em Proceedings of the 2019 Conference on Empirical Methods in
  Natural Language Processing and the 9th International Joint Conference on
  Natural Language Processing (EMNLP-IJCNLP)}, page 3982. Association for
  Computational Linguistics, 2019.

\bibitem{vit}
Alexey Dosovitskiy, Lucas Beyer, Alexander Kolesnikov, Dirk Weissenborn,
  Xiaohua Zhai, Thomas Unterthiner, Mostafa Dehghani, Matthias Minderer, Georg
  Heigold, Sylvain Gelly, et~al.
\newblock An image is worth 16x16 words.
\newblock {\em International Conference on Learning Representations (ICLR
  2021)}, 2021.

\bibitem{lora}
Edward~J Hu, Phillip Wallis, Zeyuan Allen-Zhu, Yuanzhi Li, Shean Wang, Lu~Wang,
  Weizhu Chen, et~al.
\newblock Lora: Low-rank adaptation of large language models.
\newblock In {\em International Conference on Learning Representations}, 2022.

\bibitem{graphtransformer}
Chengxuan Ying, Tianle Cai, Shengjie Luo, Shuxin Zheng, Guolin Ke, Di~He,
  Yanming Shen, and Tie-Yan Liu.
\newblock Do transformers really perform badly for graph representation?
\newblock {\em Advances in neural information processing systems},
  34:28877--28888, 2021.

\bibitem{swintransformer}
Ze~Liu, Yutong Lin, Yue Cao, Han Hu, Yixuan Wei, Zheng Zhang, Stephen Lin, and
  Baining Guo.
\newblock Swin transformer: Hierarchical vision transformer using shifted
  windows.
\newblock In {\em Proceedings of the IEEE/CVF international conference on
  computer vision}, pages 10012--10022, 2021.

\bibitem{mega}
Xuezhe Ma, Chunting Zhou, Xiang Kong, Junxian He, Liangke Gui, Graham Neubig,
  Jonathan May, and Luke Zettlemoyer.
\newblock Mega: Moving average equipped gated attention.
\newblock {\em arXiv preprint arXiv:2209.10655}, 2022.

\end{thebibliography}

\appendix
\section{Model Settings}
Actual prompt to generate bibliographic information is shown in Fig. \ref{fig:prompt}. To strictly generate structured records, we include JSON schema and "no additional explanation" descriptions.
\begin{figure*}[tbh]
\begin{center}
\begin{lstlisting}
### Instruction:
You are an academic assistant that outputs structured bibliographic data in JSON format.
Please suggest 5 recent academic papers related to "{domain}".

Each paper should be represented as a JSON object following this schema:
{
  "author": "Author name(s) in APA style, e.g., 'Smith, J. & Tanaka, K.'",
  "year": 2023,
  "title": "Title of the paper",
  "journal": "Name of the academic journal",
  "volume": "12",
  "number": "3",
  "pages": "123--145"
}

Output must be a single valid JSON array of objects and contain **no additional explanation**.
If you are unsure about any field, please leave it as an empty string ("").
\end{lstlisting}
\caption{Prompt to generate bibliographic information}\label{fig:prompt}
\end{center}
\end{figure*}

\section{Generated Records}
Some records generated with GPT-4.1 are shown in Table \ref{tab:records}. For the partially hallucinated records, the tendency for the first author, year and paper title to be strictly reproduced, while other information is reproduced in a plausible manner.
\begin{table*}[tbh]
   \caption{Examples of generated bibliographic records. Hallucinated parts are shown in bold.}\label{tab:records}
   \scalebox{0.98}{
   \begin{tabular}{p{1.5cm} p{1.5cm} p{1.1cm} p{10.2cm}}
\toprule 
\ccol{1}{Class} & \ccol{1}{Domain} & \ccol{1}{Cos} & \ccol{1}{Results}\\
\midrule
\small Completely correct & \small Vision \hspace{1cm} transformer &\ccol{1}{\small 0.995}& \small Liu, Z., Lin, Y., Cao, Y., Hu, H., Wei, Y., Zhang, Z., Lin, S., \& Guo, B. (2021). Swin Transformer: Hierarchical Vision Transformer using Shifted Windows. Proceedings of the IEEE/CVF International Conference on Computer Vision, 10012–10022.\cite{swintransformer} \\
\midrule
\small Partially \hspace{1cm} hallucinated & \small Transformer &\ccol{1}{\small 0.798}& \small Ma, X., \textbf{Xu, J., Sun, Y., He, Y., \& Lin, J. } (2022) Mega: Moving average equipped gated attention. \textbf{International Conference on Machine Learning, 162, 15369–15384}. \cite{mega} \\
\midrule
\small Completely hallucinated & \small Tabular \hspace{1cm} transformer &\ccol{1}{\small n.a.}& \small Kossen, J., Probst, P., Schirrmeister, R. T. \& Bischl, B. (2023) Self-Attention for Raw Numerical Tabular Data. IEEE Transactions on Neural Networks and Learning Systems. \\
\bottomrule
   \end{tabular}
}
\end{table*}

\end{document}